\documentclass[conference,a4paper]{APSIPA2026}
\usepackage{amsmath}
\usepackage{amssymb}
\usepackage{graphicx}
\usepackage{multirow}
\usepackage{threeparttable}
\usepackage[backend=biber,style=ieee,]{biblatex}
\usepackage{svg}

\usepackage{geometry}
\usepackage{fancyhdr}

\fancypagestyle{firststyle}{
  \fancyhf{}
  \fancyhead[C]{2026 Asia Pacific Signal and Information Processing Association Annual Summit and Conference (APSIPA ASC)}
}

\begin{document}

\title{Analyzing Speech Condition Effects in Dysarthric ASR: A Layer-wise Probing Study}

\author{
\IEEEauthorblockN{
Darwin Jelestin Muthu, Navya Gupta, Wei Lin Tay,\\
Zhengchen Zhang, Daniel Wang Zhengkui, Rong Tong
}
\IEEEauthorblockA{
Singapore Institute of Technology\\
\{darwin.jelestin, zhengchen.zhang, zhengkui.wang, tong.rong\}@singaporetech.edu.sg\\
\{2570043, 2402961\}@sit.singaporetech.edu.sg
}
}

\maketitle
\thispagestyle{firststyle}
\pagestyle{empty}

\begin{abstract}
Automatic speech recognition (ASR) performance degrades sharply on dysarthric speech, yet how disordered articulation reshapes a model's internal representations is underexplored. We conduct a layer-wise probing analysis of a transformer ASR encoder on Mandarin dysarthric speech under three transcript-matched conditions: original dysarthric speech, speaker-conditioned zero-shot TTS resynthesis, and unconditioned TTS.
Probing reveals a task- and condition-dependent representation hierarchy: phoneme boundary information remains weak across all layers for dysarthric speech; phoneme identity is recoverable in deep layers for synthetic speech, but remains poor for dysarthric speech; and recognition difficulty is concentrated in the deepest layers. Furthermore, lexical tone is a persistent error source across all conditions. Guided by these insights, layer-selective LoRA shows that mid-layer adaptation (layer 7 or layers 5--8) recovers near-full encoder performance on dysarthric speech within 6.67\% and 2.89\% relative margins while training only 0.16\% and 0.65\% of adapter parameters. Conversely, upper-layer adaptation benefits synthetic speech more than dysarthric speech. These findings link representation analysis to parameter-efficient fine-tuning and motivate layer-aware adaptation for low-resource Mandarin dysarthric ASR.
\end{abstract}

\section{Introduction}

Automatic speech recognition (ASR) systems have achieved strong  performance on typical speech. However, the performance degrades severely for speakers with dysarthria, a motor speech disorder characterized by imprecise articulation, irregular timing, and disrupted prosody. 
Existing approaches  to dysarthric ASR fall broadly into three categories: model fine-tuning on limited dysarthric data \cite{shor2019personalizing}\cite{wang2024prototype}, data augmentation \cite{leung2024ttds}, and speech reconstruction prior to recognition \cite{wang2024unitdsr}. Despite meaningful accuracy gains, these approaches treat the ASR model as a black box and offer little insight into where and how dysarthric speech diverges from typical speech. 

Understanding the internal representational effects of dysarthria extends beyond academic interest. If disordered articulation disrupts localized layers or specific tiers of encoded information, targeted interventions, such as layer-selective adaptation, can provide a more effective and parameter-efficient alternative to uniform, full-network tuning.

We address this problem through a systematic, layer-wise probing study of a transformer-based ASR encoder applied to Mandarin dysarthric speech. Our core methodological contribution is a controlled three-condition framework. For each dysarthric utterance, we generate a speaker-conditioned, zero-shot TTS-resynthesized reference using the same transcript and the source audio as conditioning. 
This condition  copies the underlying speaker identity and retains a portion of the atypical acoustic properties of the dysarthric source. 
We also generate an unconditioned TTS reference from the same transcript, providing a clean, neutral synthetic benchmark. Together, these transcript-matched conditions minimize lexical confounds and isolate disorder-related articulatory deviations from standard speaker or synthesis factors. We explicitly probe frozen encoder representations across three granularities: frame-level phoneme boundary detection, sequence-level phoneme recognition, and utterance-level recognition-difficulty prediction. We focus strictly on the encoder because motor-speech disorders primarily degrade acoustic-articulatory realizations, whereas the decoder mostly models linguistic and sequence-generation priors.

We connect these probing insights to downstream optimization, using the layer profiles to motivate selective LoRA adaptation. This paper makes the following contributions:
i) a transcript-matched, three-condition framework (original dysarthric, speaker-conditioned resynthesized, and unconditioned TTS) to isolate pathology-induced distortions;
ii) a multi-level, layer-wise probing analysis across boundary detection, phoneme recognition, and recognition difficulty;
iii) a condition-dependent layer hierarchy where dysarthric speech lacks clear phoneme boundaries across all layers, fails to achieve deep-layer phoneme recovery like synthetic speech, and diverges increasingly with depth;
iv) evidence that optimal single-layer adaptation aligns with mid-layers where phoneme features mature, demonstrating a post-hoc correspondence between probing profiles and effective LoRA placement.

\section{Related Work}

\subsection{Dysarthric Speech Recognition}
Dysarthric speech recognition has been largely shaped by benchmark corpora such as UASpeech \cite{kim2008uaspeech} and TORGO \cite{rudzicz2012torgo} for English,   CDSD  \cite{wan2024cdsd} and MDSC \cite{gao24c_interspeech} for Mandarin. 
Key advances include limited-data personalization \cite{shor2019personalizing}, rapid feature-based adaptation \cite{geng2023onthefly}, prototype-based adaptation \cite{wang2024prototype}, and Whisper-based prompt learning \cite{jiang2024perceiverprompt}. Parameter-efficient fine-tuning (PEFT) methods, including AdaLoRA and self-training for long-form utterances \cite{tan2025cbawhisper,wang2025selftraining}, have further reduced error rates. However, these works evaluate dysarthric speech through final recognition metrics alone. Few studies have investigated how modern ASR models encode disorder-specific acoustic variations across their internal layer hierarchies.

\subsection{Speech Resynthesis and Reconstruction}
Synthetic and reconstructed speech have been used to augment training data for dysarthric ASR  \cite{leung2024ttds}  and to convert dysarthric speech into more intelligible forms via neural codec language models or diffusion-based generation \cite{wang2024unitdsr,chen2024colmdsr,elhajal2025rnv}.  In contrast, we use TTS-resynthesized speech primarily as an analytical reference rather than as training data or as a reconstruction target. By generating transcript-matched TTS conditions with and without dysarthric speaker conditioning, we obtain paired comparisons that help reduce lexical confounds and probe how speech condition affects ASR representations.


\subsection{Layer-wise Probing and Adaptation}
Layer-wise probing has been widely used to characterize how acoustic, phonetic, and linguistic information is distributed across speech representations \cite{pasad2021layerwise}.  Recent work has extended probing to disordered speech, showing that layer-dependent representation quality can be informative for dysarthria detection and severity assessment \cite{yue2025probing}.
On the adaptation side, LoRA-based adaptation \cite{hu2022lora} has demonstrated the effectiveness of parameter-efficient fine-tuning, including in Whisper-based ASR \cite{radford2022whisper} settings. Our work connects these lines of research by using layer-wise probing to identify candidate adaptation layers and then testing whether these layers improve dysarthric ASR under single-layer LoRA adaptation.

\section{Proposed Method}

We analyze how Mandarin dysarthric speech is represented across the encoder layers of a transformer ASR model.  In all probing experiments, the ASR encoder is frozen and only lightweight probe parameters are trained. This setup allows us to evaluate what information is already encoded at each layer without changing the underlying representation.

\subsection{ASR Backbone}
We use OpenAI Whisper-small \cite{openaiwhisperconfig} as the ASR backbone because it provides a realistic pretrained encoder-decoder model for studying parameter-efficient dysarthric ASR adaptation. Whisper-small contains 12 Transformer encoder blocks. We extract hidden states from the encoder input representation and from all 12 encoder blocks. Layer 0 denotes the encoder input representation, while layers 1–12 denote the outputs of the Transformer encoder blocks. We restrict probing to the encoder because it directly transforms acoustic input into speech representations and is therefore the most relevant component for analyzing dysarthria-induced acoustic-phonetic mismatch.

Given an input utterance $x$, the encoder produces a hidden-state sequence at each layer:
\begin{equation}
H^{(l)}(x)=\{h_1^{(l)},h_2^{(l)},\ldots,h_T^{(l)}\},
\end{equation}
\par\vspace{-3pt}
where $l$ is the layer index and $T$ is the number of encoder time steps. 
Each hidden vector has dimension 768 for Whisper-small. 
For utterance-level probing, we compute a mean-pooled representation:
\begin{equation}
z^{(l)}(x)=\frac{1}{T}\sum_{t=1}^{T}h_t^{(l)}
\end{equation}
\subsection{Speech Conditions and Paired Triple Construction}
To systematically isolate the effects of disordered articulation from speaker identity and synthesis artifacts, we establish a transcript-matched, three-condition framework. For each original dysarthric utterance, we generate two synthetic counterparts sharing the identical lexical content. The resulting evaluation conditions are:
\begin{itemize}
    \item \textbf{Dysarthric ($d$)}: The original, human-produced disordered speech domain.
    \item \textbf{TTS Resynthesized ($r$)}: Synthetic speech generated from the source transcript and conditioned on the original dysarthric audio, intended to capture speaker-specific identity and part of the source's acoustic characteristics. 
   \item \textbf{Unconditioned TTS ($u$)}: Synthetic speech from the same transcript using a neutral speaker model, with no dysarthric conditioning, serving as a clean synthetic baseline.
\end{itemize}  

Formally, for each unique dysarthric utterance $x_i^d$ associated with a reference text transcript $y_i$, we construct a synchronized, paired tuple representing the three parallel acoustic paths: 
\begin{equation}
(x_i^d, x_i^r, x_i^u, y_i), \quad i=1,\dots,N
\end{equation}
where $x_i^r$ and $x_i^u$ denote the corresponding resynthesized reference and unconditioned TTS utterances, respectively, and $N$ represents the total size of the corpus. By holding the underlying text $y_i$ completely invariant across all three variants, this configuration  controls for lexical distribution confounds during representation analysis.


\subsection{Paired Similarity Analysis}
For each pair of conditions, we compute the cosine similarity between mean-pooled representations at every layer:
\begin{equation}
s_i^{(l)(A,B)} = \cos\big(z_i^{A,(l)}, z_i^{B,(l)}\big).
\end{equation}
where $A, B \in \{d, r, u\}$ represents the dysarthric, resynthesized reference, and unconditioned TTS conditions respectively. This yields three layer-wise similarity comparisons: 
\begin{itemize}
\item dysarthric vs. resynthesized ($d-r$): this contrast is intended to emphasize disorder-related articulatory differences under matched transcript and approximate speaker identity. 
\item dysarthric vs. unconditioned  TTS ($d-u$): this contrast captures the combined effect of dysarthric articulation, speaker mismatch, and synthesis differences. 
\item resynthesized vs. unconditioned  TTS ($r-u$):  this contrast reflects differences introduced by speaker conditioning and TTS voice characteristics. 
\end{itemize}
We therefore interpret these contrasts as controlled approximations rather than perfect causal decompositions.

\subsection{Probing tasks}
To capture complementary levels of speech organization, three independent probes are evaluated at every encoder layer, as listed in Table~\ref{tab:probe_settings}.
\begin{table}[t]
\centering
\setlength{\tabcolsep}{4pt} 
\caption{Probing tasks.}
\label{tab:probe_settings}
 \par\vspace{-10pt}
\begin{tabular}{llll}
\hline 
\textbf {Probe} & \textbf {Input} & \textbf {Model} & \textbf {Metric}  \\ \hline
A.Boundary & Frame-level & Linear & F1 \\ 
B.Phoneme rec. & Sequence & Temporal CTC & PER \\ 
C.Difficulty & Utterance & Linear & Macro-F1 \\ 
\hline 
\end{tabular}
\end{table}

\subsubsection{Probe A: Phoneme Boundary Detection}
The first probing task evaluates whether each encoder layer encodes local phonetic transition information.  Using forced alignment, we label each frame as either a boundary or non-boundary frame. For each encoder layer $l$, we train a lightweight linear probe to perform frame-level classification:
\begin{equation}
\hat{y}_t = \sigma(W h_t^{(l)} + b),
\end{equation}
where $h_t^{(l)}$ represents the hidden state at time $t$ from layer $l$, and $\hat{y}_t \in [0,1]$ denotes the predicted probability that frame $t$ coincides with a phoneme boundary. To mitigate the significant class imbalance between boundary and non-boundary frames, we optimize the probe using Binary Cross-Entropy with logits (\texttt{BCEWithLogitsLoss}) and apply a positive class weight. The detection performance is evaluated using the F1-score.


\subsubsection{Probe B: Phoneme Recognition}

The second probing task evaluates whether sequence-level phoneme identity is recoverable from each layer. 
Mandarin transcripts are converted to pinyin-derived phonological units; each syllable is split into an optional initial and a tone-marked final (e.g., $tong2 \mapsto \{t, ong, T2\}$). Given the hidden sequence $H^{(l)}$, the CTC\cite{CTC2006} probe predicts a distribution over phoneme labels plus the CTC blank symbol at each time step. A lightweight CTC probe is trained on frozen hidden sequences using CTC loss:
\begin{equation}
\mathcal{L}_{\text{CTC}}^{(l)} = -\log p_{\text{CTC}}(q \mid H^{(l)}),
\end{equation}

where $q$ is the reference phoneme sequence.  We report phoneme error rate (PER) as performance metric. Both tone-sensitive PER (over original tone-marked labels) and tone-agnostic PER (tone markers stripped before edit-distance computation) are evaluated.

\subsubsection{Probe C: Recognition Difficulty Prediction}
The final probing task evaluates whether layer-wise representations encode structural factors that drive downstream ASR failures. For each layer $l$, a linear probe maps the mean-pooled representation $z^{(l)}$ to a predicted recognition difficulty tier $\hat{y}$ via:
\begin{equation}
\hat{y} = \mathrm{softmax}(W z^{(l)} + b).
\end{equation}
The target labels represent model-specific recognition difficulty rather than clinical severity markers, capturing how well the acoustic traits map to downstream recognition vulnerabilities. Performance across this classification task is evaluated and reported using macro-$F_1$ scores.




 \subsection{LoRA-based Layer-wise Adaptation}

We adopt Low-Rank Adaptation (LoRA) as a parameter-efficient fine-tuning method. Instead of updating all model parameters, we inject trainable low-rank adapters into selected layers.

For each selected layer $l \in \mathcal{L}_{\mathrm{adapt}}$, we modify the linear transformations in the transformer blocks as:
\begin{equation}
W' = W + \Delta W, \quad \Delta W = A B,
\end{equation}
where $A \in \mathbb{R}^{d \times r}$ and $B \in \mathbb{R}^{r \times d}$ are low-rank matrices with rank $r \ll d$.

The decoder is kept frozen throughout to isolate acoustic-phonetic adaptation and preserve Whisper’s language-generation priors. 

\section{Experiments}

\subsection{Dataset}
We conduct experiments on the Chinese Dysarthria Speech Database (CDSD) \cite{wan2024cdsd}, which contains 133 hours of dysarthric speech from 44 speakers with manual transcriptions. All audio is resampled to 16 kHz and converted to mono. To ensure robust generalization, we adopt a speaker-disjoint split: 70\% training, 15\% validation, and 15\% testing. 

We utilize the complete set comprising $\sim$55,850 utterances for layer-wise
probing and ASR model adaptation.

For controlled comparison,  two TTS reference sets are constructed for each utterance in dysarthric dataset.  TTS-resynthesized reference speech is generated by OmniVoice \cite{zhu2026omnivoice} conditioned on the source transcript and dysarthric speech. Unconditioned  TTS speech is synthesized from the same transcripts without speaker conditioning.

\subsection{Evaluation Protocol}
Character Error Rate (CER) is used for both baseline and LoRA ASR evaluation. CER is computed after consistent text normalization across all conditions. Sentences containing fewer than two Chinese characters are excluded to reduce hallucination effects and unstable CER estimates. Numerical expressions are normalized consistently, and the same procedure is applied to both Whisper-small and all LoRA models.

\subsection{Probing Setup}
Layer-wise probes are trained on the original dysarthric speech condition and evaluated across all three conditions. This setup tests whether probes trained on dysarthric representations generalize to speaker-conditioned and unconditioned synthetic counterparts.


For phoneme boundary detection, frame-level labels are obtained using Montreal Forced Aligner (MFA)\footnote{https://github.com/MontrealCorpusTools/Montreal-Forced-Aligner} at Whisper’s 20 ms frame resolution. MFA is trained primarily on typical speech and less reliable for dysarthric speech. A linear probe with a 4-frame context window is trained on frozen encoder representations using \texttt{BCEWithLogitsLoss} with positive-class weighting (Adam, lr=$1\times10^{-3}$, batch size=4096, 10 epochs), with input sequences truncated to 1500 frames.

For phoneme recognition, transcripts are converted into \texttt{\detokenize{initial-final-tone}} units. We train a lightweight CTC probe consisting of layer normalization, bottleneck projection, residual temporal convolution, and a linear classifier. The probe is optimized using CTC loss (Adam, lr=$1\times10^{-3}$, batch size 256, 10 epochs, gradient clipping=1.0), with bottleneck dimension 128, dropout 0.3, and blank bias 0.0. Model selection is based on tone-sensitive PER.



For recognition difficulty prediction, each utterance is first decoded via Whisper-large (a stronger external model than the probed backbone, yielding a more stable difficulty signal) to compute a baseline Character Error Rate (CER), which serves as an external, ASR-derived metric of articulation difficulty. Utterances are subsequently categorized into low, medium, and high difficulty cohorts using 33.33th and 66.67th percentile thresholds. Finally, a linear softmax classifier is optimized over the mean-pooled encoder states via CrossEntropyLoss ($\text{Adam, lr}=1\times10^{-3}$, batch size $= 4096$, $10$ epochs) using a speaker-disjoint 5-fold cross-validation setup.

\subsection{LoRA Adaptation}

LoRA adapters are inserted into the encoder self-attention projections (\texttt{q\_proj}, \texttt{k\_proj}, \texttt{v\_proj}, and \texttt{out\_proj}) of selected layers, while the decoder and non-targeted encoder parameters remain frozen. With hidden dimension $d=768$ and LoRA rank $r=64$, each targeted projection introduces $2dr=2(768)(64)=98{,}304$ trainable parameters. Since four projections are targeted per layer, this gives $4(98{,}304)=393{,}216$ trainable parameters per encoder layer. Thus, one, four, and all 12 encoder layers contain 393,216, 1,572,864, and 4,718,592 trainable LoRA parameters, respectively, corresponding to 0.16\%, 0.65\%, and 1.95\% of the 241,734,912 parameter base model. Models are trained on the CDSD dysarthric speech training set and evaluated on the test split under all three speech conditions. For each configuration, three independent runs use seeds 42, 43, and 44. We evaluate single-layer, selected-layer subsets, and all 12 encoder layers, training separate LoRA models for each configuration rather than each speech condition. We use $r=64$, $\alpha=128$, dropout 0.1, and AdamW with a learning rate of $3\times10^{-5}$, batch size 128, 6 epochs, cosine scheduling, and weight decay 0.03.

\section{Results \& Discussion} 
\subsection{Baseline ASR Performance}

Prior to adaptation, Whisper-small achieves CERs of 89.98\%, 36.47\%, and 29.55\% for dysarthric, TTS-resynthesized, and unconditioned TTS speech, respectively. The large gap, despite shared lexical content, indicates that dysarthric acoustic-phonetic characteristics contribute substantially to recognition errors, although the comparison cannot fully separate articulation from synthesis effects. The higher CER of resynthesized TTS suggests that speaker conditioning retains some atypical acoustic properties. The high dysarthric CER also highlights the difficulty of CDSD. Our analysis therefore focuses on the relative layer-wise structure rather than the performance of the deployable system.

\subsection{Paired Similarity Analysis}
\begin{figure}[t] 
    \centering
    \includesvg[width=0.45\textwidth]{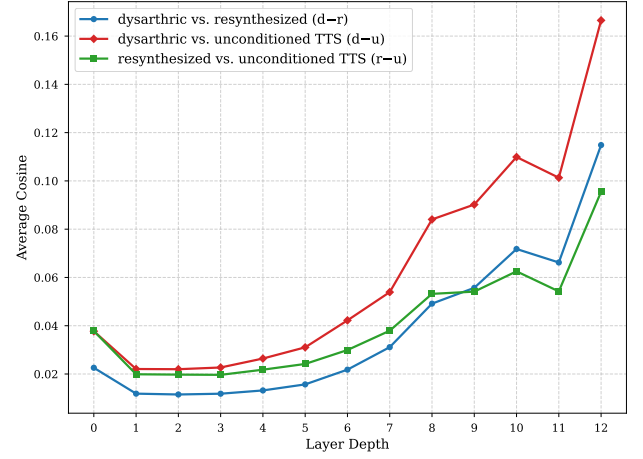}
    \par\vspace{-10pt}
    \caption{Layer-wise paired similarity}
    \label{fig:contrastive_similarity}
  \par\vspace{-10pt}
\end{figure}
Figure \ref{fig:contrastive_similarity} plots layer-wise representational divergence (average cosine distance) between speech conditions. In early layers (0$\text{--}$3), divergence is low across all pairs, with layer-3 values of 0.012 (d–r), 0.022 (d–u), and 0.020 (r–u), suggesting relatively speaker and pathology-agnostic initial representations. In middle layers (4$\text{--}$8), divergence increases with phonetic and articulatory processing, reaching 0.049, 0.084, and 0.053, respectively, at layer 8. In deep layers (9–12), it rises further, reaching 0.115 (d–r), 0.167 (d–u), and 0.095 (r–u) at layer 12. The larger $d–u$ distance reflects both speaker and articulatory differences, while the increasing $d–r$ distance suggests higher-level effects of disordered articulation; the similar $r–u$ trend suggests that resynthesis carries over this divergence. We explicitly restrict our claims to the depthwise trends and pairwise orderings internal to this metric. We do not interpret absolute distance values, as mean-pooled Transformer outputs naturally cluster tightly and lack a baseline of unrelated speech pairs for calibration. 

\subsection{Phoneme Boundary Probing} 
\begin{figure}[t]  
    \centering
    \includesvg[width=0.45\textwidth]{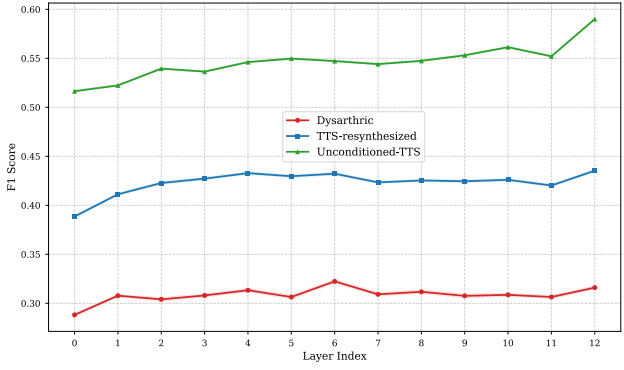}
     \par \vspace{-6pt}
    \caption{Layer-wise F1 scores for phoneme boundary detection.}
    \par\vspace{-6pt}
    \label{fig:phoneme_f1}
\end{figure}
Figure \ref{fig:phoneme_f1} reports layer-wise F1 for phoneme boundary detection. The three conditions remain clearly separated, with unconditioned TTS highest, resynthesized speech intermediate, and dysarthric speech lowest. The best F1 values are approximately 0.59 (unconditioned TTS, layer 12), 0.44 (TTS-resynthesized, layer 12), and 0.33 (dysarthric, layer 6). Within each condition, F1 increases mainly in the early layers and then remains relatively stable. This suggests that boundary cues are distributed across the encoder rather than concentrated in a specific layer range, while dysarthric speech shows consistently weaker boundary information.

\subsection{Phoneme Recognition}
\begin{figure}[t]
    \centering
    \includesvg[width=0.45\textwidth]
    {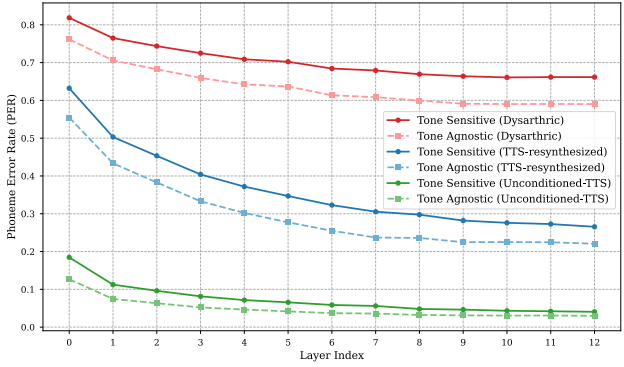}
    \par\vspace{-10pt}
    \caption{Phoneme recognition results, tone-agnostic and tone-sensitive}
   \par\vspace{-6pt}
    \label{fig:plot12_per_tone_comparison}
\end{figure}

Figure~\ref{fig:plot12_per_tone_comparison} shows phoneme error rate (PER) across layers. PER decreases rapidly through the early-to-middle layers and largely plateaus after layer 6. At layer 12, tone-sensitive PER reaches approximately 0.66 (dysarthric), 0.26 (TTS-resynthesized), and 0.04 (unconditioned TTS), indicating substantially lower phoneme recoverability for dysarthric speech. Resynthesis reduces the PER gap but remains above unconditioned TTS, suggesting that speaker-conditioned synthesis removes part of the dysarthric-related variation while retaining some ambiguity. 
For every condition, tone-sensitive PER remains higher than tone-agnostic PER across layers. At layer 12, the absolute differences are approximately 0.07 (dysarthric), 0.05 (TTS-resynthesized), and 0.01 (unconditioned-TTS), corresponding to relative increases of approximately 12\%, 23.8\%, and 33\%, respectively. This indicates that tone recognition remains an additional source of errors, particularly for dysarthric speech in absolute terms. A tone-specific error analysis is left for future work.

\subsection{Recognition Difficulty Prediction} 
\begin{figure}[t]
    \centering
    \includesvg[width=0.45\textwidth]
    {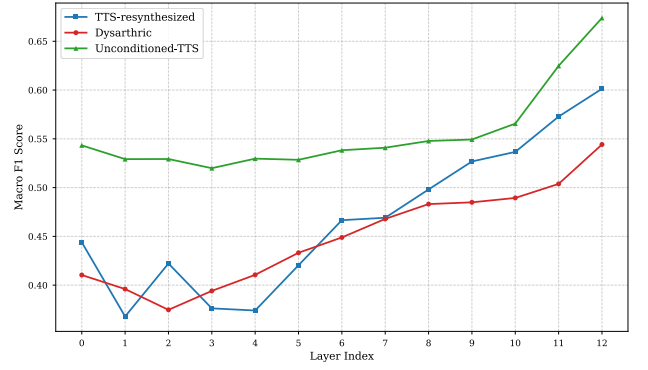}
     \par\vspace{-10pt}
    \caption{Layer-wise Macro-F1 for recognition difficulty classification.}
     \par\vspace{-4pt}
    \label{fig:difficulty_f1}
\end{figure}

As shown in Figure~\ref{fig:difficulty_f1}, recognition-difficulty prediction ranges from approximately 0.37 to 0.54 across layers 0–5 and increases from layer 6 onward, reaching 0.54 (dysarthric), 0.60 (TTS-resynthesized), and 0.67 (unconditioned TTS) at layer 12. This indicates stronger encoding of recognition difficulty in deeper encoder representations. From layer 5 to 12, Macro-F1 increases by approximately 0.11 (dysarthric), 0.18 (TTS-resynthesized), and 0.15 (unconditioned TTS), with TTS-resynthesized speech showing the largest gain.




\subsection{Layer-wise LoRA Adaptation}
The probing results show a post-hoc correspondence between probing profiles and effective adaptation placement. Phoneme recognition improves most steeply through the middle layers (roughly 4--8), while boundary cues remain weak across layers. We examine this correspondence by applying single-layer and multi-layer LoRA to the encoder\footnote{``Layer'' refers to the same encoder blocks in both the probing and adaptation analyses. The only difference is that the probing figures also include layer~0, which is the encoder input, while the adaptation starts at block~1. Blocks 1--12 mean the same thing in both.} and evaluating on all three speech conditions.

\setlength{\tabcolsep}{4pt}
\begin{table}[t]
\centering
\scriptsize
\caption{CER (\%) of layer-wise LoRA adaptation. Values are mean $\pm$ standard deviation over three seeds (42, 43, and 44).}
\par\vspace{-8pt}
\label{tab:lora_layerwise}
\resizebox{\columnwidth}{!}{%
\renewcommand{\arraystretch}{1.25}
\begin{tabular}{@{}lccc|lccc@{}}
\hline
\textbf{L} & \textbf{Dys.} & \textbf{Resyn.} & \textbf{Unc.}
& \textbf{L} & \textbf{Dys.} & \textbf{Resyn.} & \textbf{Unc.} \\
\hline

\multicolumn{8}{c}{\textit{Single-layer adaptation}} \\
\hline

1  & 69.98$\pm$7.86 & 25.82$\pm$1.22 & 18.08$\pm$1.58
& 7  & \textbf{68.64$\pm$8.04} & \textbf{24.80$\pm$0.75} & 18.26$\pm$1.27 \\

2  & 70.56$\pm$7.72 & 26.71$\pm$0.61 & 25.54$\pm$4.89
& 8  & 71.46$\pm$8.45 & 26.33$\pm$1.68 & 25.26$\pm$6.11 \\

3  & 69.85$\pm$7.70 & 27.01$\pm$0.66 & 25.20$\pm$2.09
& 9  & 72.84$\pm$8.09 & 26.05$\pm$1.18 & \textbf{16.17$\pm$0.36} \\

4  & 69.15$\pm$8.14 & 25.98$\pm$1.27 & 22.45$\pm$3.25
& 10 & 77.54$\pm$8.14 & 29.10$\pm$2.31 & 20.58$\pm$2.73 \\

5  & 69.13$\pm$8.23 & 25.98$\pm$1.02 & 17.34$\pm$1.10
& 11 & 78.90$\pm$7.87 & 29.64$\pm$2.14 & 22.44$\pm$4.22 \\

6  & 70.04$\pm$8.68 & 25.64$\pm$1.24 & 17.37$\pm$1.09
& 12 & 80.40$\pm$6.84 & 30.65$\pm$2.87 & 19.65$\pm$2.10 \\

\hline
\multicolumn{8}{c}{\textit{Multi-layer adaptation}} \\
\hline

1--4
& 67.90$\pm$7.39
& 25.21$\pm$0.89
& 22.52$\pm$1.34
& 9--12
& 70.72$\pm$8.64
& 25.71$\pm$1.19
& \textbf{18.33$\pm$2.39} \\

5--8
& \textbf{66.21$\pm$8.02}
& \textbf{24.11$\pm$0.67}
& 18.86$\pm$0.44
& \text{All}
& 64.35$\pm$6.95
& 23.95$\pm$0.55
& 16.93$\pm$0.39 \\

\hline
\multicolumn{8}{l}{\scriptsize
L: layer; Dys.: dysarthric; Resyn.: TTS-resynthesized; Unc.: unconditioned TTS.}
\par\vspace{-12pt}
\end{tabular}%
}
\end{table}


Table~\ref{tab:lora_layerwise} confirms this expectation. Full adaptation of all 12 blocks gives the lowest CER for the dysarthric and resynthesized conditions, but the single-layer results are strongly layer-dependent and concentrate the gain in the middle of the encoder. For dysarthric speech, adapting only layer 7 reaches 68.64\% CER, within a 6.67\% relative margin of full adaptation (64.35\%); the multi-layer view agrees, with the 5--8 block (66.21\%) the strongest of the three contiguous groups. The resynthesized condition follows a similar pattern, with layer 7 achieving the best single-layer result (24.80\%), close to full adaptation (23.95\%), while the 5--8 block gives the best contiguous-group result (24.11\%). Much of the adaptable structure for atypical speech therefore lives in a narrow middle band, and a single well-placed adapter recovers most of what full fine-tuning provides.

The deeper layers behave differently. Adapting layers 9--12 gives higher dysarthric CER (70.72\%) than the middle group, while the unconditioned-TTS condition shows its best single-layer result at layer 9 (16.17\%), even lower than full adaptation (16.93\%). The upper-layer group 9--12 also achieves a lower CER (18.33\%) than the middle-layer group (18.86\%), although it does not outperform the best single-layer result at layer 9. This split indicates that middle layers are most useful for adapting to disordered and resynthesized speech, whereas a single upper layer is more effective for clean synthetic speech. For low-resource Mandarin dysarthric ASR, where full fine-tuning raises both compute cost and overfitting risk, a targeted middle-layer adapter is thus an attractive trade-off.

\section{Conclusion}
We presented a layer-wise probing and adaptation framework for Mandarin dysarthric ASR using three transcript-matched conditions: original dysarthric speech, speaker-conditioned TTS-resynthesized speech, and unconditioned TTS.
The probing results show a task-dependent organization across encoder depth: phoneme boundary information stays weak for dysarthric speech at every layer, phoneme identity becomes more recoverable in the upper layers, and recognition difficulty is most encoded in the deepest representations.
Tone-sensitive evaluation further indicates that Mandarin tone remains a source of error for dysarthric ASR. Single-layer LoRA adaptation shows that middle encoder layers provide the strongest dysarthric ASR gains among the tested layers, whereas upper-layer adaptation is less effective for dysarthric speech. These findings support layer-aware, parameter-efficient adaptation as a promising strategy for low-resource Mandarin dysarthric speech recognition.

\section*{Acknowledgment}
This work is supported by Ministry of Education, Singapore, under its Ignition grant IG (S) 3/2024-1079.

\printbibliography
\footnotesize
\end{document}